%% file: 00_attention_might.tex
\documentclass[11pt,a4paper]{article}
\usepackage[hyperref]{emnlp-ijcnlp-2019}
\usepackage{times}
\usepackage{latexsym}
\usepackage{url}

\usepackage{ucs}
\usepackage{fontenc}
\usepackage[utf8x]{inputenc}
\usepackage[english]{babel}
\usepackage{dsfont}
\usepackage{textcomp}
\usepackage{amsfonts}
\usepackage{amsmath}
\usepackage{amssymb}
\usepackage{graphicx}
\usepackage{booktabs}
\usepackage{multirow}

\usepackage{color}
\usepackage{tcolorbox}
\usepackage{CJK}
\usepackage{adjustbox}
\tcbset{width=0.9\textwidth,boxrule=0pt,colback=red,arc=0pt,auto outer arc,left=0pt,right=0pt,boxsep=5pt}
\usepackage{transparent}

\usepackage{tikz}
\usetikzlibrary{
    decorations.pathreplacing,
    calligraphy,
    shapes,
    shapes.geometric,
    shapes.symbols,
    shapes.arrows,
    shapes.multipart,
    shapes.callouts,
    shapes.misc}
\pgfdeclarelayer{bg}
\pgfsetlayers{bg,main}

\DeclareMathOperator*{\KL}{\textsc{kl}}
\DeclareMathOperator*{\JSD}{\textsc{jsd}}
\DeclareMathOperator*{\TVD}{\textsc{tvd}}

\aclfinalcopy 

\title{Attention is not not Explanation}

\author{Sarah Wiegreffe\thanks{\enspace Equal contributions.} \\
  School of Interactive Computing \\
  Georgia Institute of Technology \\
  {\tt saw@gatech.edu} \\\And
  Yuval Pinter\footnotemark[1] \\
  School of Interactive Computing \\
  Georgia Institute of Technology \\
  {\tt uvp@gatech.edu}\\}

\date{}

\begin{document}
\maketitle

\begin{abstract}
  Attention mechanisms play a central role in NLP systems, especially within recurrent neural network (RNN) models.
  Recently, there has been increasing interest in whether or not the intermediate representations offered by these modules may be used to explain the reasoning for a model's prediction, and consequently reach insights regarding the model's decision-making process.
  A recent paper claims that `Attention is not Explanation' \cite{jain2019attention}.
  We challenge many of the assumptions underlying this work, arguing that such a claim depends on one's definition of explanation, and that testing it needs to take into account all elements of the model.
  We propose four alternative tests to determine when/whether attention can be used as explanation: a simple uniform-weights baseline; a variance calibration based on multiple random seed runs; a diagnostic framework using frozen weights from pretrained models; and an end-to-end adversarial attention training protocol.
  Each allows for meaningful interpretation of attention mechanisms in RNN models.
  We show that even when reliable adversarial distributions can be found, they don't perform well on the simple diagnostic, indicating that prior work does not disprove the usefulness of attention mechanisms for explainability.
\end{abstract}

\input{01_intro}

\input{02a_thesis}

\input{02b_antithesis}

\input{03_randomness}

\input{04_alternative}

\input{05_background}

\input{06_summary}

\bibliography{attention_might}
\bibliographystyle{acl_natbib}

\newpage

\appendix

\input{90_stats}

\input{92_all_advs}

\input{94_class_plots}

\end{document}

%% file: 01_intro.tex
\section{Introduction}

\input{fig_roadmap}

Attention mechanisms \cite{bahdanau2014neural} are nowadays ubiquitous in NLP, and their suitability for providing explanations for model predictions is a topic of high interest \cite{xu2015show,rocktaschel2015reasoning, mullenbach2018explainable, thorne2019generating, serrano2019attention}.
If they indeed offer such insights, many application areas would benefit by better understanding the internals of neural models that use attention as a means for, e.g., model debugging or architecture selection.
A recent paper \cite{jain2019attention} points to possible pitfalls that may cause researchers to misapply attention scores as explanations of model behavior, based on a premise that explainable attention distributions should be \emph{consistent} with other feature-importance measures as well as \emph{exclusive} given a prediction.\footnote{A preliminary version of our theoretical argumentation was published as a blog post on Medium at \url{http://bit.ly/2OTzU4r}. 
Following the ensuing online discussion,
the authors uploaded a post-conference version of the paper to \href{https://arxiv.org/abs/1902.10186}{arXiv} (\textsc{v3}) which addresses some of the issues in the post.
We henceforth refer to this later version.}
Its core argument, which we elaborate in \S \ref{sec:bg}, is that if alternative attention distributions exist that produce similar results to those obtained by the original model, then the original model's attention scores cannot be reliably used to ``faithfully'' explain the model's prediction. 
Empirically, the authors show that achieving such alternative distributions is easy for a large sample of English-language datasets.

We contend (\S \ref{sec:anti}) that while \citeauthor{jain2019attention} ask an important question, and raise a genuine concern regarding potential misuse of attention weights in explaining model decisions on English-language datasets, some key assumptions used in their experimental design leave an implausibly large amount of freedom in the setup, 
ultimately leaving practitioners without an applicable way for measuring the utility of attention distributions in specific settings.

We apply a more model-driven approach to this question, beginning (\S \ref{ssec:unif}) with testing attention modules' \textbf{contribution} to a model by applying a simple baseline where attention weights are frozen to a uniform distribution.
We demonstrate that for some datasets, a frozen attention distribution performs just as well as learned attention weights, concluding that randomly- or adversarially-perturbed distributions are not evidence against attention as explanation in these cases.
We next (\S \ref{ssec:seed}) examine the \textbf{expected variance} in attention-produced weights by initializing multiple training sequences with different random seeds, allowing a better quantification of how much variance can be expected in trained models.
We show that considering this background stochastic variation when comparing adversarial results with a traditional model allows us to better interpret adversarial results.
In \S \ref{ssec:guide}, we present a simple yet effective \textbf{diagnostic tool} which tests attention distributions for their usefulness by using them as frozen weights in a non-contextual multi-layered perceptron (MLP) architecture.
The favorable performance of LSTM-trained weights provides additional support for the coherence of trained attention scores. This demonstrates a sense in which attention components indeed provide a meaningful model-agnostic interpretation of tokens in an instance.

In \S \ref{sec:oppose}, we introduce a \textbf{model-consistent} training protocol for finding adversarial attention weights, correcting some flaws we found in the previous approach. We train a model using a modified loss function which takes into account the distance from an ordinarily-trained base model's attention scores in order to learn parameters for adversarial attention distributions.
We believe these experiments are now able to support or refute a claim of faithful explainability, by providing a way for convincingly saying by construction that a plausible alternative `explanation' can (or cannot) be constructed for a given dataset and model architecture.
We find that while plausibly adversarial distributions of the consistent kind can indeed be found for the binary classification datasets in question, they are not as extreme as those found in the inconsistent manner, as illustrated by an example from the \textsc{IMDb} task in Figure \ref{tab:concrete}.
Furthermore, these outputs do not fare well in the diagnostic MLP, calling into question the extent to which we can treat them as equally powerful for explainability.

Finally, we provide a theoretical discussion (\S \ref{sec:defns}) on the definitions of interpretability and explainability, grounding our findings within the accepted definitions of these concepts.

Our four quantitative experiments are illustrated in \autoref{fig:roadmap}, where each bracket on the left covers the components in a standard RNN-with-attention architecture which we manipulate in each experiment.
We urge NLP researchers to consider applying the techniques presented here on their models containing attention in order to evaluate its effectiveness at providing explanation. We offer our code for this purpose at \url{https://github.com/sarahwie/attention}.

\input{tab_concrete}

%% file: fig_roadmap.tex
\begin{figure*}
    \centering
    
    \begin{tikzpicture}[
      hid/.style 2 args={
        rectangle split,
        rectangle split horizontal,
        draw=#2,
        rectangle split parts=#1,
        fill=#2!50,
        outer sep=0.6mm},
      vec/.style 2 args={
        rectangle split,
        draw=#2,
        rectangle split parts=#1,
        fill=#2!50,
        outer sep=0.6mm}]
      \fontfamily{epigrafica}
      \small
      \foreach \i [count=\step from 1] in {the,movie,was,good}
        \node (i\step) at (2*\step, -5) {\normalsize\emph\i};
      \node[hid={1}{gray}] (o0) at (5.25, 0) {}; 
      \node[vec={3}{orange}] (p) at (9.5, -2) {};
      \foreach \step in {1,...,4} {
        \node[hid={1}{teal}] (a\step) at (2*\step+0.75, -1) {};
        \node[hid={3    }{red}] (h\step) at (2*\step, -3) {};
        \node[hid={4}{blue}] (e\step) at (2*\step, -4) {};
        \begin{pgfonlayer}{bg}
            \draw[->] (i\step.north) -> (e\step.south);
            \draw[->] (e\step.north) -> (h\step.south);
            \draw[->] (h\step.north) -> (a\step.south west);
            \draw[->] (h\step.north) -> (o0.south);
            \draw[->] (a\step.north) -> (o0.south);
            \draw[->] (p.west) -> (a\step.south east);
        \end{pgfonlayer}
      }
      \foreach \step in {1,...,3} {
        \pgfmathtruncatemacro{\next}{add(\step,1)}
        \draw[<->] (h\step.east) -> (h\next.west);
      }
      \node[anchor=west] at (-1, 0) {Prediction Score};
      \node[anchor=west] at (-1, -1) {Attention Scores};
      \node[anchor=west] at (-1, -2) {Attention Parameters};
      \node[anchor=west] at (-1, -3) {LSTM};
      \node[anchor=west] at (-1, -4) {Embedding};
      \draw[decoration={calligraphic brace}, decorate, thick, anchor=east] (-1.3,-1.4) node {} -- (-1.3,-0.6);
      \draw[decoration={calligraphic brace}, decorate, thick, anchor=east] (-2.3,-1.4) node {} -- (-2.3,0.4);
      \node[draw,fill=white] at (-2.3,-2) {$\emptyset$};
      \draw[decoration={calligraphic brace}, decorate, thick, anchor=east] (-2.3,-4.4) node {} -- (-2.3,-2.6);
      \draw[decoration={calligraphic brace}, decorate, thick, anchor=east] (-3.3,-4.4) node {} -- (-3.3,0.4);
      \draw[decoration={calligraphic brace}, decorate, thick, anchor=east] (-4.3,-1.4) node {} -- (-4.3,0.4);
      \node[draw,fill=white] at (-4.4,-3) {MLP};
      \begin{pgfonlayer}{bg}
        \draw[decoration={calligraphic brace}, decorate, thick, anchor=east] (-4.3,-4.4) node {} -- (-4.3,-2.6);
      \end{pgfonlayer}
      \draw[decoration={calligraphic brace}, decorate, thick, anchor=east] (-5.3,-4.4) node {} -- (-5.3,-0.6);
      \normalsize
      \node at (-1.3, 1.0) {J\&W};
      \node at (-2.3, 1.0) {\S \ref{ssec:unif}};
      \node at (-3.3, 1.0) {\S \ref{ssec:seed}};
      \node at (-4.3, 1.0) {\S \ref{ssec:guide}};
      \node at (-5.3, 1.0) {\S \ref{sec:oppose}};
    \end{tikzpicture}
    
    \caption{Schematic diagram of a classification LSTM model with attention, including the components manipulated or replaced in the experiments performed in \newcite{jain2019attention} and in this work (by section).
    }
    \label{fig:roadmap}
\end{figure*}
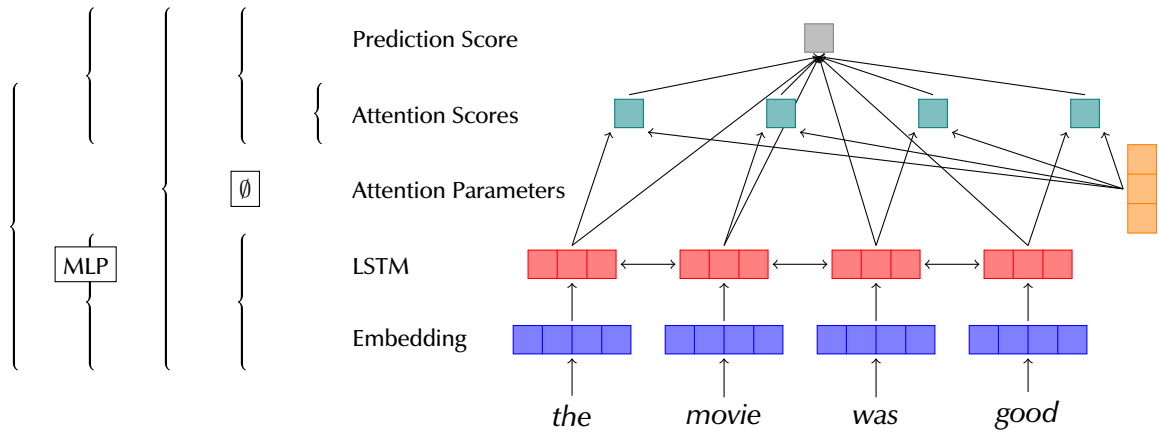

%% file: tab_concrete.tex
\begin{table*}
    \centering
    \begin{tabular}{ll}
        Base model & \input{heatmaps/gold.tex} \\
        \citet{jain2019attention} & \input{heatmaps/jw.tex} \\
        Our adversary & \input{heatmaps/our.tex}
    \end{tabular}
\captionof{figure}{Attention maps for an IMDb instance (all predicted as positive with score $> 0.998$), showing that in practice it is difficult to learn a distant adversary which is consistent on all instances in the training set.\label{tab:concrete}}
\end{table*}

%% file: heatmaps/gold.tex
{\transparent{0.0}\colorbox{red}{\transparent{1.0}{{\strut brilliant}}}}
{\transparent{0.3559606373310089}\colorbox{red}{\transparent{1.0}{{\strut and}}}}
{\transparent{0.05411434918642044}\colorbox{red}{\transparent{1.0}{{\strut moving}}}}
{\transparent{0.06780438125133514}\colorbox{red}{\transparent{1.0}{{\strut performances}}}}
{\transparent{0.1428971290588379}\colorbox{red}{\transparent{1.0}{{\strut by}}}}
{\transparent{0.005474465433508158}\colorbox{red}{\transparent{1.0}{{\strut tom}}}}
{\transparent{0.009820275940001011}\colorbox{red}{\transparent{1.0}{{\strut and}}}}
{\transparent{0.03517894074320793}\colorbox{red}{\transparent{1.0}{{\strut peter}}}}
{\transparent{0.16998445987701416}\colorbox{red}{\transparent{1.0}{{\strut finch}}}}
 
 

%% file: heatmaps/jw.tex
{\transparent{0.0}\colorbox{blue}{\transparent{1.0}{{\strut brilliant}}}}
{\transparent{0.00010042172652902082}\colorbox{blue}{\transparent{1.0}{{\strut and}}}}
{\transparent{0.0}\colorbox{blue}{\transparent{1.0}{{\strut moving}}}}
{\transparent{0.9971297383308411}\colorbox{blue}{\transparent{1.0}{{\strut \textcolor{white}{performances}}}}}
{\transparent{0.0002744705998338759}\colorbox{blue}{\transparent{1.0}{{\strut by}}}}
{\transparent{0.0}\colorbox{blue}{\transparent{1.0}{{\strut tom}}}}
{\transparent{0.0018750614253804088}\colorbox{blue}{\transparent{1.0}{{\strut and}}}}
{\transparent{0.00029212282970547676}\colorbox{blue}{\transparent{1.0}{{\strut peter}}}}
{\transparent{0.0}\colorbox{blue}{\transparent{1.0}{{\strut finch}}}}
 
 

%% file: heatmaps/our.tex
{\transparent{0.0}\colorbox{green}{\transparent{1.0}{{\strut brilliant}}}}
{\transparent{0.19920063018798828}\colorbox{green}{\transparent{1.0}{{\strut and}}}}
{\transparent{0.21356631815433502}\colorbox{green}{\transparent{1.0}{{\strut moving}}}}
{\transparent{0.14148923754692078}\colorbox{green}{\transparent{1.0}{{\strut performances}}}}
{\transparent{0.15089991688728333}\colorbox{green}{\transparent{1.0}{{\strut by}}}}
{\transparent{0.1841667741537094}\colorbox{green}{\transparent{1.0}{{\strut tom}}}}
{\transparent{0.00011674504639813676}\colorbox{green}{\transparent{1.0}{{\strut and}}}}
{\transparent{0.10151346772909164}\colorbox{green}{\transparent{1.0}{{\strut peter}}}}
{\transparent{0.0}\colorbox{green}{\transparent{1.0}{{\strut finch}}}}

%% file: 02a_thesis.tex
\section{Attention Might be Explanation}
\label{sec:bg}

In this section, we briefly describe the experimental design of \newcite{jain2019attention} and look at the results they provide to support their claim that `Attention is not explanation'.
The authors select eight classification datasets, mostly binary, and two question answering tasks for their experiments (detailed in \S \ref{ssec:setup}).

They first present a correlation analysis of attention scores and other interpretability measures.
They find that attention is not strongly correlated with other, well-grounded feature importance metrics, specifically gradient-based and leave-one-out methods (which in turn correlate well with each other).
This experiment evaluates the authors' claim of \emph{consistency} -- that attention-based methods of explainability cannot be valid if they do not correlate well with other metrics.
We find the experiments in this part of the paper convincing and do not focus our analysis here.
We offer our simple MLP diagnostic network (\S \ref{ssec:guide}) as an additional way for determining validity of attention distributions, in a more \emph{in vivo} setting.

Next, the authors present an adversarial search for alternative attention distributions which minimally change model predictions.
To this end, they manipulate the attention distributions of trained models (which we will call \textbf{base} from now on) to discern whether alternative distributions exist for which the model outputs near-identical prediction scores.
They are able to find such distributions, first by randomly permuting the base attention distributions on the test data during model inference, and later by adversarially searching for maximally different distributions that still produce a prediction score within $\epsilon$ of the base distribution.
They use these experimental results as supporting evidence for the claim that attention distributions cannot be explainable because they are not \emph{exclusive}.
As stated, the lack of comparable change in prediction with a change in attention scores is taken as evidence for a lack of ``faithful'' explainability of the attention mechanism from inputs to output.

Notably, \citeauthor{jain2019attention} detach the attention distribution and output layer of their pretrained network from the parameters that compute them (see \autoref{fig:roadmap}), treating each attention score as a standalone unit independent of the model.
In addition, they compute an independent adversarial distribution for each instance.

%% file: 02b_antithesis.tex
\subsection{Main Claim}
\label{sec:anti}

We argue that \citeauthor{jain2019attention}'s counterfactual attention weight experiments do not advance their thesis, for the following reasons:

\paragraph{Attention Distribution is not a Primitive.}
From a modeling perspective, detaching the attention scores obtained by parts of the model (i.e. the attention mechanism) degrades the model itself.
The base attention weights are not assigned arbitrarily by the model, but rather computed by an integral component whose parameters were trained alongside the rest of the layers;
the way they work depends on each other.
\citeauthor{jain2019attention} provide alternative distributions which may result in similar predictions, but in the process they remove the very linkage which motivates the original claim of attention distribution explainability, namely the fact that the model was \emph{trained} to attend to the tokens it chose.
A reliable adversary must take this consideration into account, as our setup in \S \ref{sec:oppose} does.

\paragraph{Existence does not Entail Exclusivity.}
On a more theoretical level, we hold that attention scores are used as providing \emph{an} explanation; not \emph{the} explanation.
The final layer of an LSTM model may easily produce outputs capable of being aggregated into the same prediction in various ways, however the model still makes the choice of a specific weighting distribution using its trained attention component.
This mathematically flexible production capacity is particularly evident in binary classifiers, where prediction is reduced to a single scalar, and an average instance (of e.g. the \textsc{IMDb} dataset) might contain 179 tokens, i.e. 179 scalars to be aggregated.
This effect is greatly exacerbated when performed independently on each instance.\footnote{Indeed, the most open-ended task, question answering over CNN data, produces considerable difficulty to manipulate its scores by random permutation (Figure 6e in \newcite{jain2019attention}).
Similarly, the adversarial examples presented in Appendix C of the paper for the QA datasets select a different token of the correct word's type, which should not surprise us even under an LSTM assumption (encoder hidden states are typically affected by the input word to a noticeable degree).}
Thus, it is no surprise that \citeauthor{jain2019attention} find what they are looking for given this degree of freedom.

In summary, due to the per-instance nature of the demonstration and the fact that model parameters have not been learned or manipulated directly, \citeauthor{jain2019attention} have \textbf{not shown the existence of an adversarial model} that produces the claimed adversarial distributions.
Thus, we cannot treat these adversarial attentions as equally plausible or faithful explanations for model prediction.
Additionally, they haven't provided a baseline of how much variation is to be expected in learned attention distributions, leaving the reader to question just how adversarial the found adversarial distributions are.

%% file: 03_randomness.tex
\section{Examining Attention Distributions}
\label{sec:rand}

In this section, we apply a careful methodological approach for examining the properties of attention distributions and propose alternatives. We begin by identifying the appropriate scope of the models' performance and variance, followed by implementing an empirical diagnostic technique which measures the model-agnostic usefulness of attention weights in capturing the relationship between inputs and output.

\subsection{Experimental Setup}
\label{ssec:setup}

\input{tab_datastats}
\input{tab_unif}

In order to make our many points in a succinct fashion as well as follow the conclusions drawn by \citeauthor{jain2019attention}, we focus on experimenting with the binary classification subset of their tasks, and on models with an LSTM architecture \cite{hochreiter1997long}, the only one the authors make firm conclusions on.
Future work may extend our experiments to extractive tasks like question answering, as well as other attention-prone tasks, like seq2seq models.

We experiment on the following datasets: Stanford Sentiment Treebank (\textsc{SST}) \cite{socher2013recursive}, \textsc{IMDb} Large Movie Reviews Corpus \cite{maas2011learning}, \textsc{20 Newsgroups} (hockey vs. baseball),\footnote{\url{http://qwone.com/~jason/20Newsgroups/}} the \textsc{AG News} Corpus,\footnote{\url{http://www.di.unipi.it/~gulli/AG_corpus_of_news_articles.html}} and two prediction tasks from \textsc{MIMIC-III ICD9} \cite{johnson2016mimic}: \textsc{Diabetes} and \textsc{Anemia}.
The tasks are as follows: to predict positive or negative sentiment from sentences (\textsc{SST}) and movie reviews (\textsc{IMDb}), to predict the topic of news articles as either baseball (neg.) or hockey (pos.) in \textsc{20 Newsgroups} and either world (neg.) or business (pos.) in \textsc{AG News}, to predict whether a patient is diagnosed with diabetes from their ICU discharge summary, and to predict whether the patient is diagnosed with acute (neg.) or chronic (pos.) anemia (both \textsc{MIMIC-III ICD9}). We use the dataset versions, including train-test split, provided by \citeauthor{jain2019attention}.\footnote{\url{https://github.com/successar/AttentionExplanation}}
All datasets are in English.\footnote{We do not include the Twitter Adverse Drug Reactions (ADR) \cite{nikfarjam2015pharmacovigilance} dataset as the source tweets are no longer all available.}
Data statistics are provided in \autoref{tab:data}.

We use a single-layer bidirectional LSTM with $\tanh$ activation, followed by an additive attention layer \cite{bahdanau2014neural} and softmax prediction, which is equivalent to the \textsc{LSTM} setup of \citeauthor{jain2019attention}.
We use the same hyperparameters found in that work to be effective in training, which we corroborated by reproducing its results to a satisfactory degree (see middle columns of \autoref{tab:unif}).
We refer to this architecture as the \textbf{main setup}, where training results in a \textbf{base model}.

Following \citeauthor{jain2019attention}, all analysis is performed on the test set.
We report F1 scores on the positive class, and apply the same metrics they use for model comparison, namely Total Variation Distance (TVD) for comparing prediction scores $\hat{y}$ and Jensen-Shannon Divergence (JSD) for comparing weighting distributions $\alpha$:

\begin{equation*}
    \TVD(\hat y_1, \hat y_2) = \frac{1}{2} \sum_{i=1}^{|\mathcal{Y}|} \left|\hat y_{1i} - \hat y_{2i}\right|;
\end{equation*}

\begin{equation*}
    \JSD(\alpha_1, \alpha_2) = \frac{1}{2} \KL[\alpha_1 \parallel \bar{\alpha}] + \frac{1}{2}\KL[\alpha_2 \parallel \bar{\alpha}],
\end{equation*}
where $\bar{\alpha} = \tfrac{\alpha_1 + \alpha_2}{2}$.

\subsection{Uniform as the Adversary}
\label{ssec:unif}

\input{fig_seed}

First, we test the validity of the classification tasks and datasets by examining whether attention is necessary in the first place.
We argue that if attention models are not useful compared to very simple baselines, i.e. their parameter capacity is not being used, there is no point in using their outcomes for any type of explanation to begin with.
We thus introduce a \textbf{uniform} model variant, identical to the main setup except that the attention distribution is frozen to uniform weights over the hidden states.

The results comparing this baseline with the base model are presented in \autoref{tab:unif}. If attention was a necessary component for good performance, we would expect a large drop between the two rightmost columns.
Somewhat surprisingly, for three of the classification tasks the attention layer appears to offer little to no improvement whatsoever.
We conclude that these datasets, notably \textsc{AG News} and \textsc{20 Newsgroups}, are not useful test cases for the debated question: attention is not explanation if you don't need it.
We subsequently ignore the two News datasets, but keep SST, which we deem borderline.

\subsection{Variance within a Model}
\label{ssec:seed}

We now test whether the variances observed by \citeauthor{jain2019attention} between trained attention scores and adversarially-obtained ones are unusual.
We do this by repeating their analysis on eight models trained from the main setup using different initialization random seeds.
The variance introduced in the attention distributions represents a baseline amount of variance that would be considered normal.

The results are plotted in \autoref{fig:seeds} using the same plane as \citeauthor{jain2019attention}'s Figure 8 (with two of these reproduced as (e-f)). Left-heavy violins are interpreted as data classes for which the compared model produces attention distributions similar to the base model, and so having an adversary that manages to `pull right' supports the argument that distributions are easy to manipulate.
We see that SST distributions (c, e) are surprisingly robust to random seed change, validating our choice to continue examining this dataset despite its borderline F1 score.
On the Diabetes dataset, the negative class is already subject to relatively arbitrary distributions from the different random seed settings (d), making the highly divergent results from the overly-flexible adversarial setup (f) seem less impressive.
Our consistently-adversarial setup in \S \ref{sec:oppose} will further explore the  difficulty of surpassing seed-induced variance between attention distributions.

\subsection{Diagnosing Attention Distributions by Guiding Simpler Models}
\label{ssec:guide}

As a more direct examination of models, and as a complementary approach to \newcite{jain2019attention}'s measurement of backward-pass gradient flows through the model for gauging token importance, we introduce a post-hoc training protocol of a non-contextual model \textbf{guided} by pre-set weight distributions.
The idea is to examine the prediction power of attention distributions in a `clean' setting, where the trained parts of the model have no access to neighboring tokens of the instance.
If pre-trained scores from an attention model perform well, we take this to mean they are helpful and consistent, fulfilling a certain sense of explainability.
In addition, this setup serves as an effective diagnostic tool for assessing the utility of adversarial attention distributions:
if such distributions are truly alternative, they should be equally useful as guides as their base equivalent, and thus perform comparably.

\input{fig_mlp_arch}

Our diagnostic model is created by replacing the main setup's LSTM and attention parameters with a token-level affine hidden layer with $\tanh$ activation (forming an MLP), and forcing its output scores to be weighted by a pre-set, per-instance distribution, during both training and testing.
This setup is illustrated in \autoref{fig:mlparch}.
The guide weights we impose are the following:
\textbf{Uniform},
where we force the MLP outputs to be considered equally across each instance, effectively forming an unweighted baseline;
\textbf{Trained MLP},
where we do not freeze the weights layer, instead allowing the MLP to learn its own attention parameters;\footnote{This is the same as \citeauthor{jain2019attention}'s \emph{average} setup.}
\textbf{Base LSTM},
where we take the weights learned by the base LSTM model's attention layer;
and \textbf{Adversary},
based on distributions found adversarially using the consistent training algorithm from \S \ref{sec:oppose} below (where their results will be discussed).

\input{tab_mlp}

The results are presented in \autoref{tab:mlp}.
The first important result, consistent across datasets, is that using pre-trained LSTM attention weights is better than letting the MLP learn them on its own, which is in turn better than the unweighted baseline.
Comparing with results from \S \ref{ssec:unif}, we see that this setup also outperforms the LSTM trained with uniform attention weights, suggesting that the attention module is more important than the word-level architecture for these datasets.
These findings strengthen the case counter to the claim that attention weights are arbitrary: independent token-level models that have no access to contextual information find them useful, indicating that they encode some measure of token importance which is not model-dependent.

%% file: tab_datastats.tex
\begin{table}
    \centering
    \small
    \begin{tabular}{lccc}
        \toprule
        Dataset & Avg. Length & Train Size & Test Size \\
         & (tokens) & (neg/pos) & (neg/pos) \\
        \midrule
        Diabetes & 1858 & 6381/1353 & 1295/319 \\
        Anemia & 2188 & 1847/3251 & 460/802 \\
        IMDb & 179 & 12500/12500 & 2184/2172 \\
        SST & 19 & 3034/3321 & 863/862 \\
        AgNews & 36 & 30000/30000 & 1900/1900 \\
        20News & 115 & 716/710 & 151/183 \\
        \bottomrule
    \end{tabular}
    \caption{Dataset statistics.}
    \label{tab:data}
\end{table}  

%% file: tab_unif.tex
\begin{table}
    \centering
    \small
    \begin{tabular}{lccc}
        \toprule
        Dataset & \multicolumn{2}{c}{Attention (Base)} & Uniform \\
         & Reported & Reproduced &  \\
        \midrule
        Diabetes & 0.79 & 0.775 & 0.706 \\
        Anemia & 0.92 & 0.938 & 0.899 \\
        IMDb & 0.88 & 0.902 & 0.879 \\
        SST & 0.81 & 0.831 & 0.822 \\
        AgNews & 0.96 & 0.964 & 0.960 \\
        20News & 0.94 & 0.942 & 0.934 \\
        \bottomrule
    \end{tabular}
    \caption{Classification F1 scores (1-class) on attention models, both as reported by \citeauthor{jain2019attention} and in our reproduction, and on models forced to use uniform attention over hidden states.}
    \label{tab:unif}
\end{table}

%% file: fig_seed.tex
\begin{figure*}
    \centering
    \small
    \begin{tabular}{cc@{\hskip 1cm}c}
        \includegraphics[width=4cm]{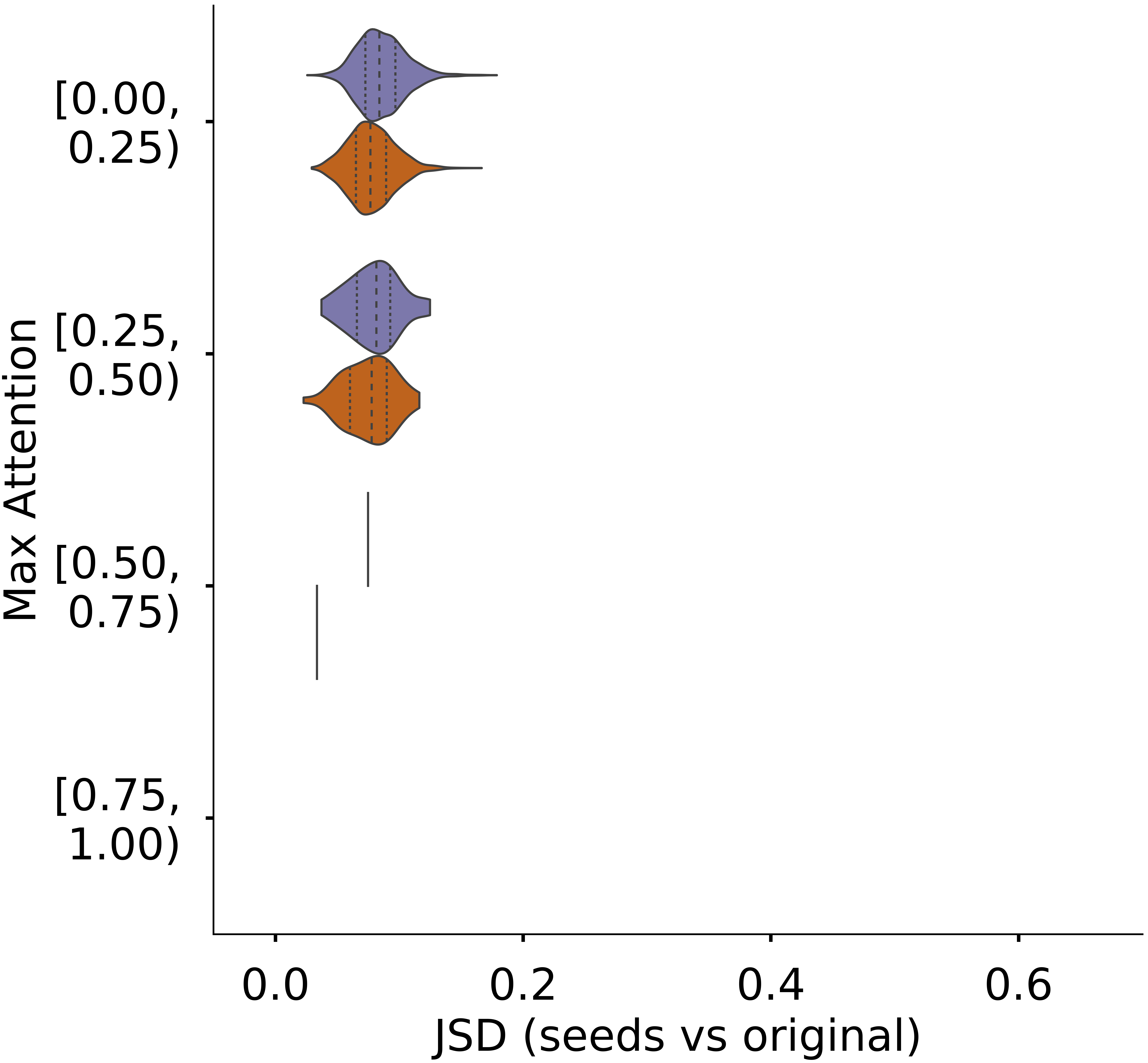}
        & 
        \includegraphics[width=4cm]{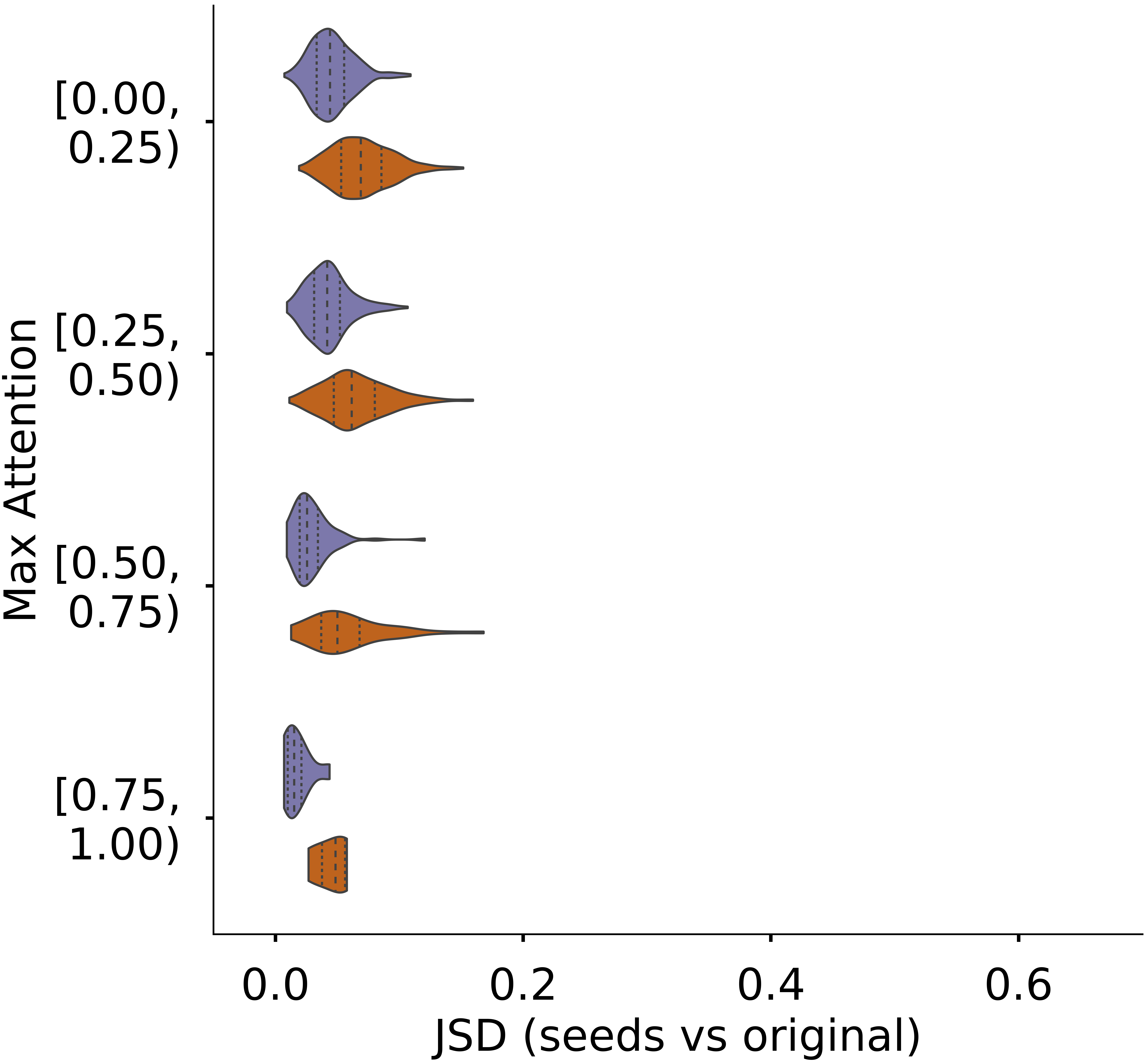}
        & 
        \includegraphics[width=4cm]{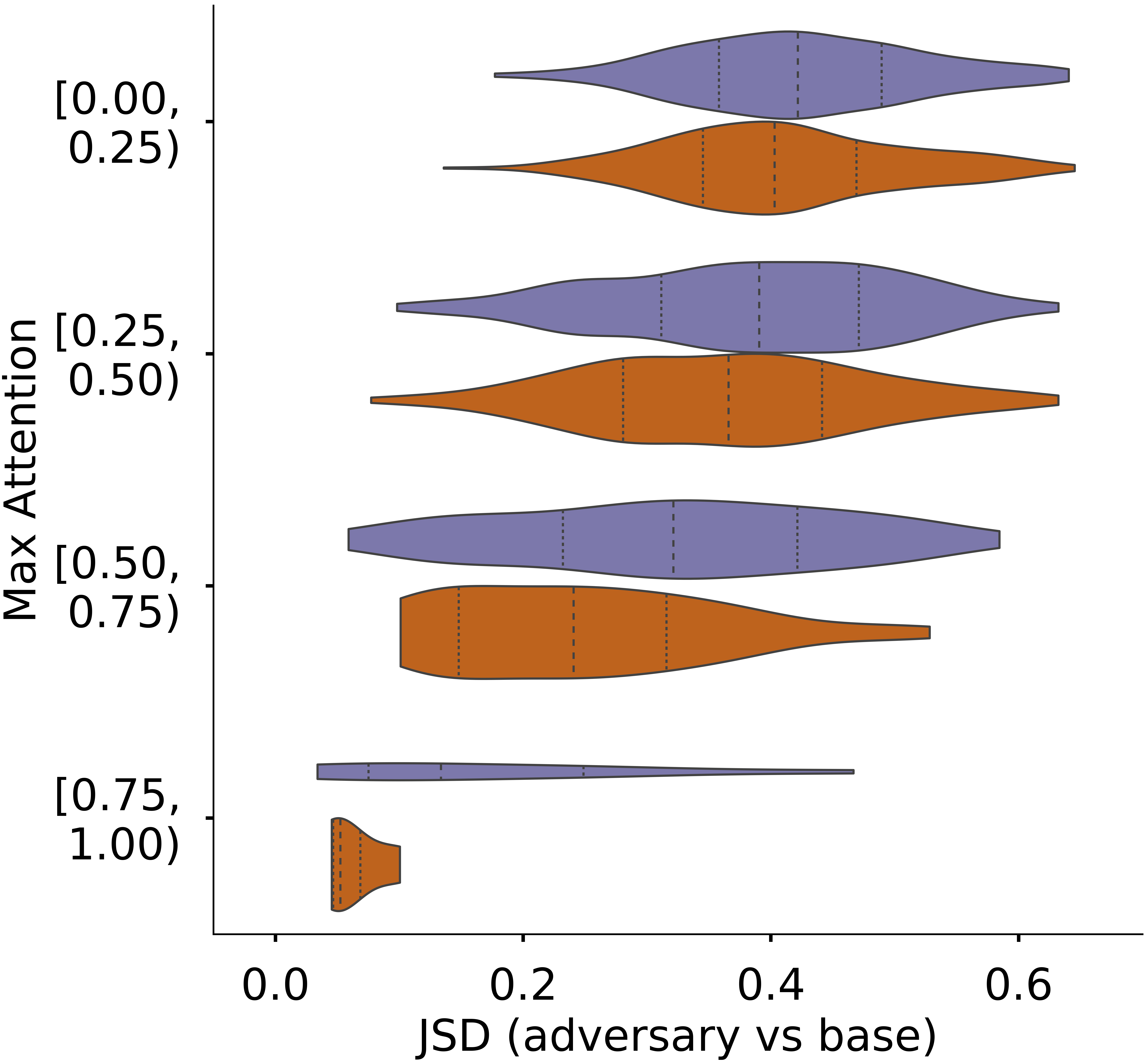} \\
        (a) \textsc{IMDb} (seeds) & (c) SST (seeds) & (e) SST (adversary) \\
        \includegraphics[width=4cm]{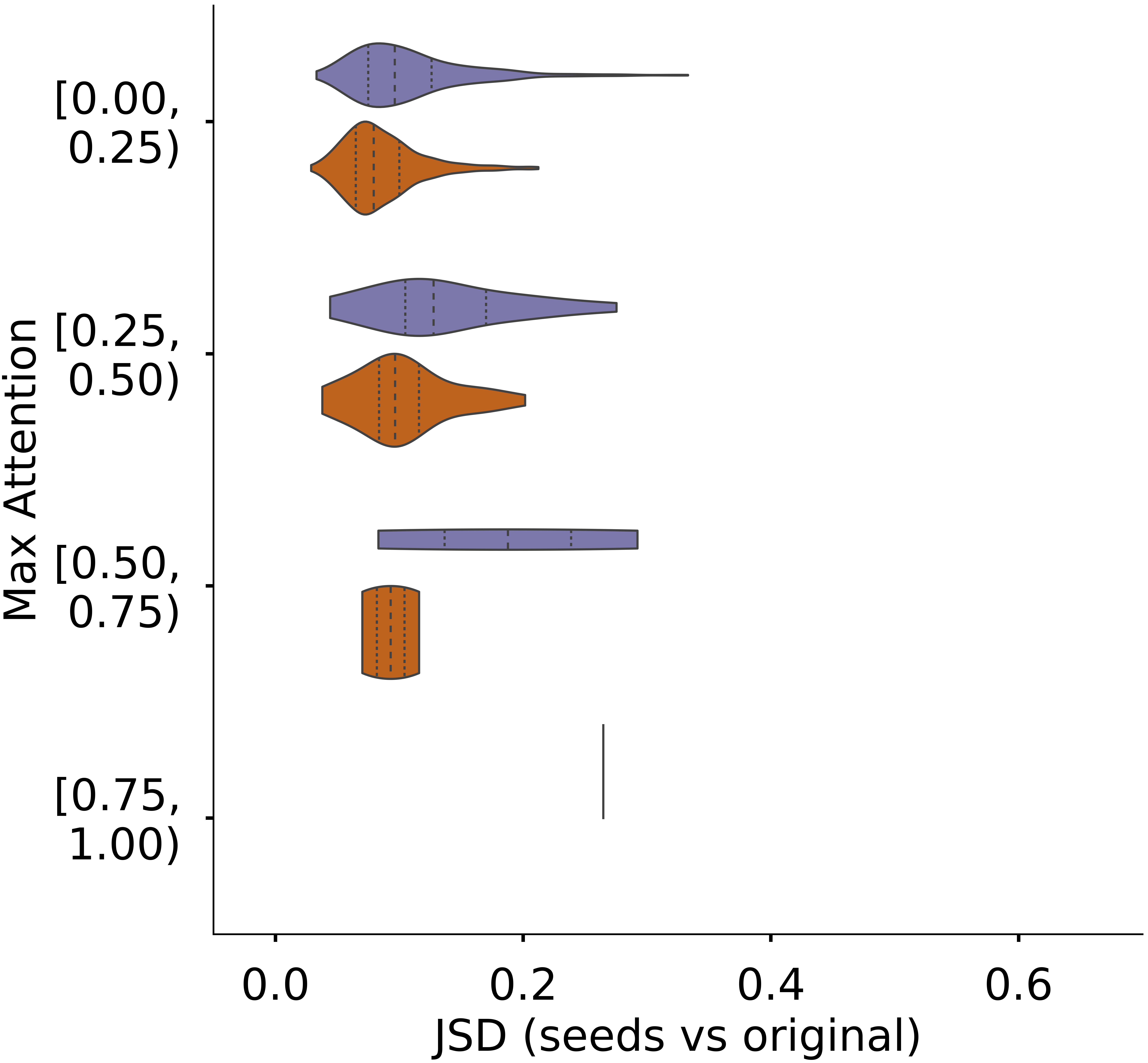}
        &
        \includegraphics[width=4cm]{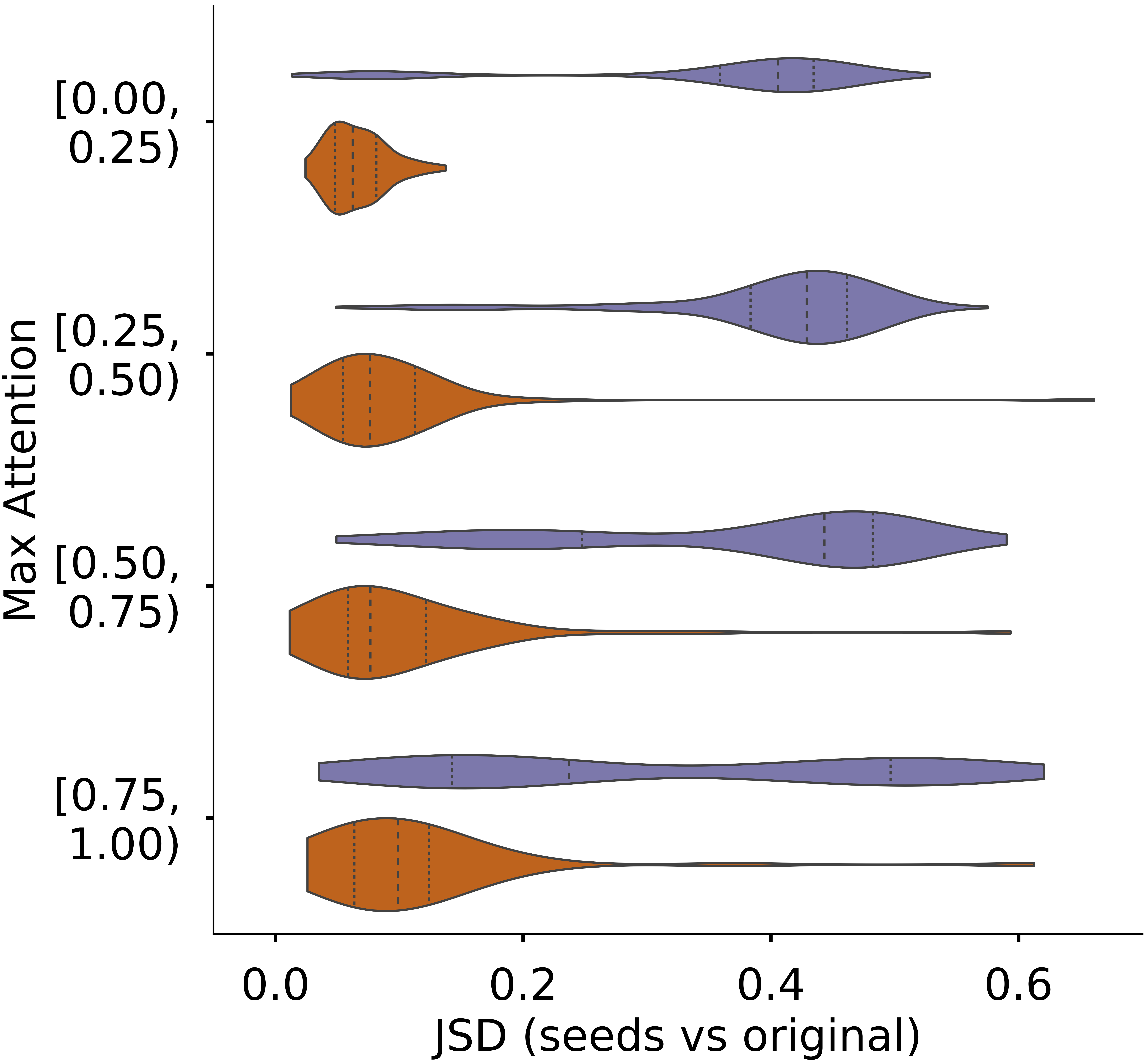}
        & 
        \includegraphics[width=4cm]{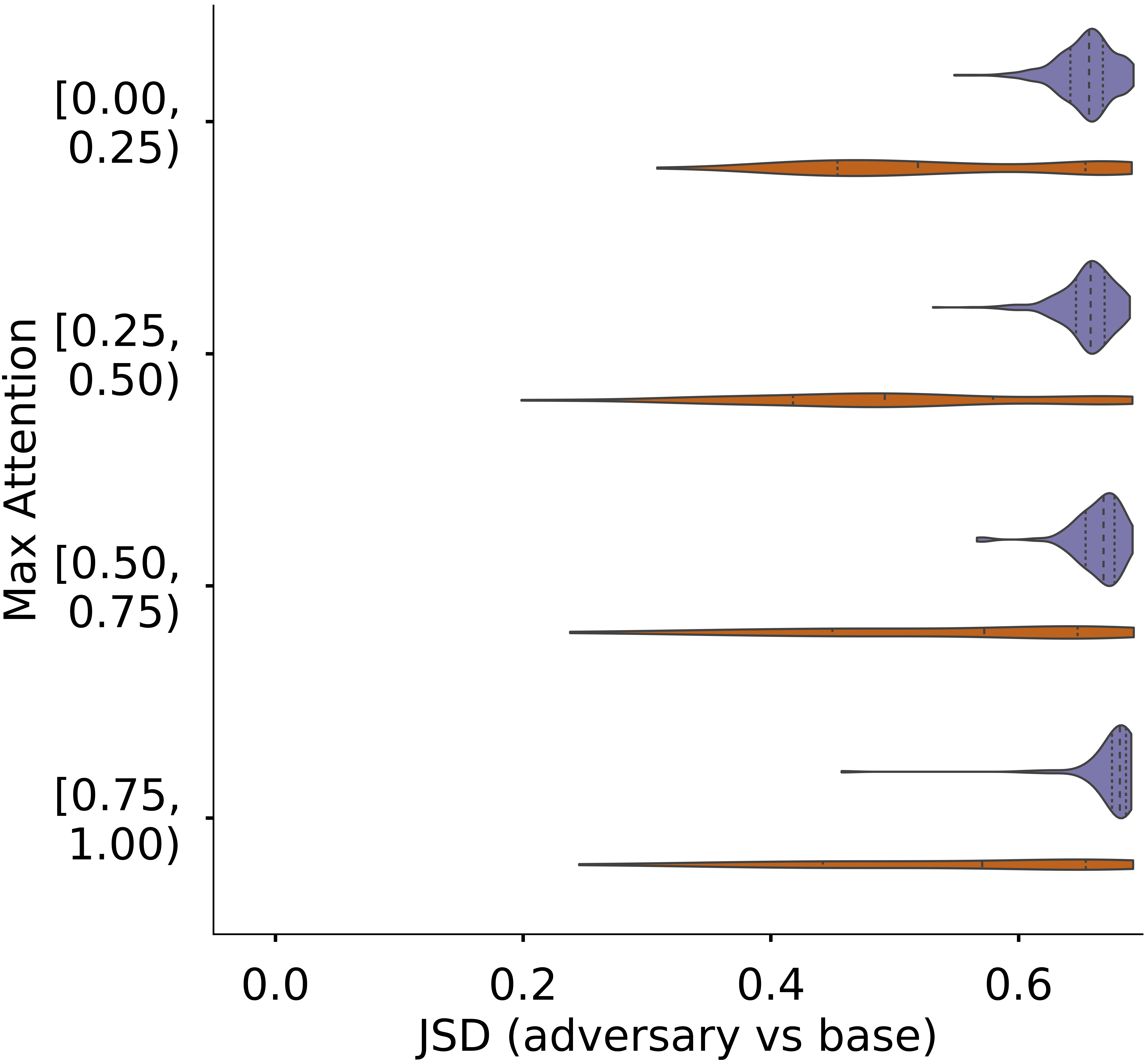}
        \\
        (b) Anemia (seeds) & (d) Diabetes (seeds) & (f) Diabetes (adversary) \\
    \end{tabular}
    \caption{Densities of maximum JS divergences (x-axis) as a function of the max attention  (y-axis) in each instance between the base distributions and: (a-d) models initialized on different random seeds; (e-f) models from a per-instance adversarial setup (replication of Figure 8a, 8c resp. in \newcite{jain2019attention}).
    In each max-attention bin, top (blue) is the negative-label instances, bottom (red) positive-label instances.
    }
    \label{fig:seeds}
\end{figure*}

%% file: fig_mlp_arch.tex
\begin{figure}
    \centering
    \small
    \begin{tikzpicture}[
      hid/.style 2 args={
        rectangle split,
        rectangle split horizontal,
        scale=0.75,
        draw=#2,
        rectangle split parts=#1,
        fill=#2!50,
        outer sep=0.6mm}]
      \fontfamily{epigrafica}
      \small
      \foreach \i [count=\step from 1] in {the,movie,was,good}
        \node (i\step) at (1.5*\step, -3.8) {\emph\i};
      \node[hid={1}{gray}] (o0) at (3.75, 0) {};
      \foreach \step in {1,...,4} {
        \node[hid={1}{black},] (a\step) at (1.5*\step, -1.2) {};
        \node[hid={3}{red}] (eh\step) at (1.5*\step, -2.1) {};
        \node[hid={4}{blue}] (e\step) at (1.5*\step, -3) {};
        \begin{pgfonlayer}{bg}
            \draw[->] (i\step.north) -> (e\step.south);
            \draw[->] (e\step.north) -> (eh\step.south);
            \draw[->] (eh\step.north) -> (o0.south);
            \draw[->, color=red] (a\step.north) -> (o0.south);
        \end{pgfonlayer}
      }
      \node[anchor=west] at (-1.0, 0) {Prediction Score};
      \node[anchor=west] at (-1.0, -1.0) {Weights};
      \node[anchor=west] at (-1.0, -1.4) {(Imposed)};
      \node[anchor=west] at (-1.0, -2.1) {Affine};
      \node[anchor=west] at (-1.0, -3) {Embedding};
    \end{tikzpicture}
    
    \caption{Diagram of the setup in \S \ref{ssec:guide} (except \textsc{Trained MLP}, which learns weight parameters).}
    \label{fig:mlparch}
\end{figure}
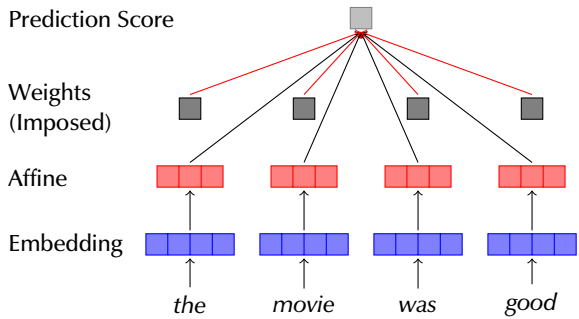

%% file: tab_mlp.tex
\begin{table}
    \centering
    \small
    \begin{tabular}{lcccc} \toprule
        Guide weights & Diab. & Anemia & SST & IMDb \\
        \midrule
        \textsc{Uniform} & 0.404 & 0.873 & 0.812 & 0.863 \\
        \textsc{Trained MLP} & 0.699 & 0.920 & 0.817 & 0.888 \\
        \textsc{Base LSTM} & \textbf{0.753} & 0.931 & \textbf{0.824} & \textbf{0.905} \\
        \textsc{Adversary} (\ref{sec:oppose}) & 0.503 & \textbf{0.932} & 0.592 & 0.700 \\
        \bottomrule
    \end{tabular}
    \caption{F1 scores on the positive class for an MLP model trained on various weighting guides.
    For \textsc{Adversary}, we set $\lambda \leftarrow 0.001$.}
    \label{tab:mlp}
\end{table}

%% file: 04_alternative.tex
\section{Training an Adversary}
\label{sec:oppose}

Having demonstrated three methods which test the meaningfulness of attention distributions as instruments of explainability with adequate control,
we now propose a model-consistent training protocol for finding adversarial attention distributions through a coherent parameterization, which holds across all training instances.
We believe this setup is able to advance the search for faithful explainability
(see \S \ref{sec:defns}).
Indeed, our results will demonstrate that the extent to which a model-consistent adversary can be found varies across datasets, and that the dramatic reduction in degree of freedom compared to previous work allows for better-informed analysis.

\paragraph{Model.}
Given the base model $\mathcal{M}_\text{b}$, we train a model $\mathcal{M}_\text{a}$ whose explicit goal is to provide similar prediction scores for each instance, while distancing its attention distributions from those of $\mathcal{M}_\text{b}$.
Formally, we train the adversarial model using stochastic gradient updates based on the following loss formula (summed over instances in the minibatch):

\begin{equation*}
    \small
    \mathcal{L}(\mathcal{M}_a, \mathcal{M}_b)^{(i)} = \TVD(\hat{y}^{(i)}_a,\hat{y}^{(i)}_b) - \lambda~ \KL(\boldsymbol{\alpha}^{\textbf{(i)}}_a \parallel \boldsymbol{\alpha}^{\textbf{(i)}}_b),
\end{equation*}
where $\hat{y}^{(i)}$ and $\boldsymbol{\alpha}^{(i)}$ denote predictions and attention distributions for an instance $i$, respectively.

$\lambda$ is a hyperparameter which we use to control the tradeoff between relaxing the prediction distance requirement (low TVD) in favor of more divergent attention distributions (high JSD), and vice versa.
When this interaction is plotted on a two-dimensional axis, the shape of the plot can be interpreted to either support the `attention is not explanation' hypothesis if it is convex (JSD is easily manipulable), or oppose it if it is concave (early increase in JSD comes at a high cost in prediction precision).

\input{fig_tvd}

\paragraph{Prediction performance.}
By definition, our loss objective does not directly consider actual prediction performance.
The TVD component pushes it towards the same score as the base model, but our setup does not ensure generalization from train to test.
It would thus be interesting to inspect the extent of the implicit F1/TVD relationship.
We report the highest F1 scores of models whose attention distributions diverge from the base, on average, by at least $0.4$ in JSD, as well as their $\lambda$ setting and corresponding comparison metrics, in \autoref{tab:ours} (full results available in \autoref{app:all_advs}).
All F1 scores are on par with the original model results reported in \autoref{tab:unif}, indicating the effectiveness of our adversarial models at imitating base model scores on the test sets.

\paragraph{Adversarial weights as guides.}
We next apply the diagnostic setup introduced in \S \ref{ssec:guide} by training a guided MLP model on the adversarially-trained attention distributions.
The results, reported in the bottom line of \autoref{tab:mlp}, show that despite their local decision-imitation abilities, they are usually completely incapable of providing a non-contextual framework with useful guides.\footnote{We note the outlying result achieved on the Anemia dataset.
This can be explained via the data distribution, which is heavily skewed towards positive examples (see \autoref{tab:data} in the Appendix), together with the fact (conceded in \newcite{jain2019attention}'s section 4.2.1) that positive instances in detection datasets such as MIMIC tend to contain a handful of indicative tokens, making the particularly helpful distributions reached by a trained model hard to replace by an adversary.
Together, this leads to the selected setting of $\lambda=0.001$ producing average distributions substantially more similar to the base than in the other datasets (JSD $\sim 0.58$ vs. $>0.61$) and thus more useful to the MLP setup.}
We offer these results as evidence that adversarial distributions, even those obtained consistently for a dataset, deprive the underlying model from some form of understanding it gained over the data, one that it was able to leverage by tuning the attention mechanism towards preferring `useful' tokens.

\input{tab_our_adv}

\paragraph{TVD/JSD tradeoff.}
In \autoref{fig:jsdtvd} we present the levels of prediction variance (TVD) allowed by models achieving increased attention distance (JSD) on all four datasets.
The convex shape of most curves does lend support to the claim that attention scores are easily manipulable;
however the extent of this effect emerging from \citeauthor{jain2019attention}'s per-instance setup is a considerable exaggeration, as seen by its position (\textcolor{red}{$\boldsymbol{+}$}) well below the curve of our parameterized model set.
Again, the SST dataset emerges as an outlier: not only can JSD be increased practically arbitrarily without incurring prediction variance cost, the uniform baseline (\textcolor{cyan}{$\blacksquare$}) comes up under the curve, i.e. with a better adversarial score. We again include random seed initializations (\textcolor{olive}{$\blacktriangle$}) in order to quantify a baseline amount of variance.

TVD/JSD plots broken down by prediction class are available in \autoref{app:class_plots}.
In future work, we intend to inspect the potential of multiple adversarial attention models existing side-by-side, all distant enough from each other.

\paragraph{Concrete Example.}

\autoref{tab:concrete} illustrates the difference between inconsistently-achieved adversarial heatmaps and consistently trained ones.
Despite both adversaries approximating the desired prediction score to very high degree, the heatmaps show that \citeauthor{jain2019attention}'s model has distributed all of the attention weight to an ad-hoc token, whereas our trained model could only distance itself from the base model distribution by so much, keeping multiple tokens in the $>0.1$ score range.

%% file: fig_tvd.tex
\begin{figure}
    \centering
    \includegraphics[width=7.7cm]{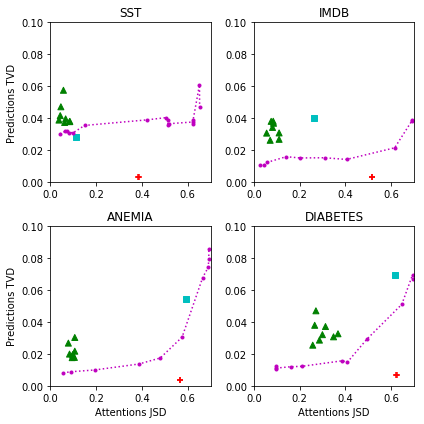}
    \caption{Averaged per-instance test set JSD and TVD from base model for each model variant. JSD is bounded at $\sim 0.693$. \textcolor{olive}{$\blacktriangle$}: random seed; \textcolor{cyan}{$\blacksquare$}: uniform weights; dotted line: our adversarial setup as $\lambda$ is varied; \textcolor{red}{$\boldsymbol{+}$}: adversarial setup from \newcite{jain2019attention}.}
    \label{fig:jsdtvd}
\end{figure}

%% file: tab_our_adv.tex
\begin{table}
    \centering
    \small
    \begin{tabular}{lcccc}
        \toprule
        Dataset & $\lambda$ & F1 ($\uparrow$) & TVD ($\downarrow$) & JSD ($\uparrow$) \\
        \midrule
        Diabetes & 2e-4 & 0.775 & 0.015 & 0.409 \\
        Anemia & 5e-4 & 0.942 & 0.017 & 0.481 \\
        SST & 5.25e-4 & 0.823 & 0.036 & 0.514 \\
        IMDb & 8e-4 & 0.906 & 0.014 & 0.405 \\
        \bottomrule
    \end{tabular}
    \caption{Best-performing adversarial models with instance-average JSD $> 0.4$.}
    \label{tab:ours}
\end{table}

%% file: 05_background.tex
\section{Defining Explanation}
\label{sec:defns}

The umbrella term of ``Explainable AI'' encompasses at least three distinct notions: \emph{transparency}, \emph{explainability}, and \emph{interpretability}.
\newcite{lipton2016mythos} categorizes transparency, or overall human understanding of a model, and post-hoc explainability as two competing notions under the umbrella of interpretability. The relevant sense of transparency, as defined by \newcite{lipton2016mythos} (\S 3.1.2), pertains to the way in which a specific portion of a model corresponds to a human-understandable construct (which \newcite{doshi2017towards} refer to as a ``cognitive chunk'').
Under this definition, it should appear sensible of the NLP community to treat attention scores as a vehicle of (partial) transparency.
Attention mechanisms do provide a look into the inner workings of a model, as they produce an easily-understandable weighting of hidden states.

\newcite{rudin2018please} defines explainability as simply a plausible (but not necessarily faithful) reconstruction of the decision-making process, and \newcite{riedl2019human} classifies explainable rationales as valuable in that they mimic what we as humans do when we rationalize past actions: we invent a story that plausibly justifies our actions, even if it is not an entirely accurate reconstruction of the neural processes that produced our behavior at the time.
Distinguishing between interpretability and explainability as two separate notions, \newcite{rudin2018please} argues that interpretability is more desirable but more difficult to achieve than explainability, because it requires presenting humans with a big-picture understanding of the correlative relationship between inputs and outputs (citing the example of linear regression coefficients). \newcite{doshi2017towards} break down interpretability into further subcategories, depending on the amount of human involvement and the difficulty of the task. 

In prior work, \newcite{lei2016rationalizing} train a model to simultaneously generate rationales and predictions from input text, using gold-label rationales to evaluate their model.
Generally, many accept the notion of extractive methods such as \newcite{lei2016rationalizing}, in which explanations come directly from the input itself (as in attention), as plausible.
Works such as \newcite{mullenbach2018explainable} and  \newcite{ehsan2019automated} use human evaluation to evaluate explanations; the former based on attention scores over the input, and the latter based on systems with additional rationale-generation capability.
The authors show that rationales generated in a post-hoc manner increase user trust in a system. 

Citing \newcite{ross2017right}, \citeauthor{jain2019attention}'s requisite for attention distributions to be used as explanation is that there must only exist one or a few closely-related correct explanations for a model prediction. However, \newcite{doshi2017towards} caution against applying evaluations and terminology broadly without clarifying task-specific explanation needs. 
If we accept the \citeauthor{rudin2018please} and \citeauthor{riedl2019human} definitions of explainability as providing a \emph{plausible}, but not necessarily \emph{faithful} rationale for model prediction, then the argument against attention mechanisms because they are not exclusive as claimed by \citeauthor{jain2019attention} is invalid, and human evaluation (which they do not consult) is necessary to evaluate the plausibility of generated rationales. Just because there exists another explanation does not mean that the one provided is false or meaningless, and under this definition the existence of multiple different explanations is not necessarily indicative of the quality of a single one.

\citeauthor{jain2019attention} define attention and explanation as measuring the ``responsibility'' each input token has on a prediction. This aligns more closely with the more rigorous
\cite[\S 3.1.1]{lipton2016mythos} 
definition of transparency, or \citet{rudin2018please}'s definition of interpretability: human understanding of the model as a whole rather than of its respective parts.
The ultimate question posed so far as `is attention explanation?' seems to be: do high attention weights on certain elements in the input lead the model to make its prediction?
This question is ultimately left largely unanswered by prior work, as we address in previous sections.
However, under the given definition of transparency, the authors' exclusivity requisite is well-defined and we find value in their counterfactual framework as a concept -- if a model is capable of producing multiple sets of diverse attention weights for the same prediction, then the relationship between inputs and outputs used to make predictions is not understood by attention analysis.
This provides us with the motivation to implement the adversarial setup coherently and to derive and present conclusions from it.
To this end, we additionally provide our \S \ref{ssec:guide} model to test the relationship between input tokens and output.

In the terminology of \newcite{doshi2017towards}, our proposed methods provide a \textit{functionally-grounded} evaluation of attention as explanation, i.e. an analysis conducted on proxy tasks without human evaluation. We believe the proxies we have provided can be used to test the validity of attention as a form of explanation from the ground-up, based on the type of explanation one is looking for.

%% file: 06_summary.tex
\section{Attention is All you Need it to Be}

Whether or not attention is explanation depends on the definition of explainability one is looking for: \emph{plausible} or \emph{faithful} explanations (or both).
We believe that prior work focused on providing plausible rationales is not invalidated by \citeauthor{jain2019attention}'s or our results.
However, we have confirmed that adversarial distributions can be found for LSTM models in some classification tasks, as originally hypothesized by \citeauthor{jain2019attention}.
This should provide pause to researchers who are looking to attention distributions for one true, faithful interpretation of the link their model has established between inputs and outputs.
At the same time, we have provided a suite of experiments that researchers can make use of in order to make informed decisions about the quality of their models' attention mechanisms when used as explanation for model predictions.

We've shown that alternative attention distributions found via adversarial training methods perform poorly relative to traditional attention mechanisms when used in our diagnostic MLP model.
These results indicate that trained attention mechanisms in RNNs on our datasets do in fact learn something meaningful about the relationship between tokens and prediction which cannot be easily `hacked' adversarially.

We view the conditions under which adversarial distributions can actually be found in practice to be an important direction for future work.
Additional future directions for this line of work include application on other tasks such as sequence modeling and multi-document analysis (NLI, QA); extension to languages other than English; and adding a human evaluation for examining the level of agreement with our measures.
We also believe our work can provide value to theoretical analysis of attention models, motivating development of analytical methods to estimate the usefulness of attention as an explanation based on dataset and model properties.

\section*{Acknowledgments}
We thank Yoav Goldberg for preliminary comments on the idea behind the original Medium post. 
We thank the online community who participated in the discussion following the post, and particularly Sarthak Jain and Byron Wallace for their active engagement, as well as for the high-quality code they released which allowed fast reproduction and modification of their experiments.
We thank Erik Wijmans for early feedback. 
We thank the members of the Computational Linguistics group at Georgia Tech for discussions and comments, particularly Jacob Eisenstein and Murali Raghu Babu.
We thank the anonymous reviewers for many useful comments.

YP is a Bloomberg Data Science PhD Fellow.

%% file: 90_stats.tex
\section{Compute and Environment}

We estimate roughly 250 model training commands were executed for this study, running for a range of 10-120 minutes each, nearly all on a single local NVIDIA Tesla K40 GPU.
The state of Georgia generates electricity from mostly Gas (43\%) and nuclear (30\%) resources;
9\% from renewables.\footnote{\url{https://www.eia.gov/state/data.php?sid=GA\#EnergyIndicators}}

%% file: 92_all_advs.tex
\section{All Results on Adversarial Setup}
\label{app:all_advs}

\begin{table*}
    \centering
    \begin{tabular}{cc}
        \toprule
        \begin{tabular}{lcccc}
            Dataset & $\lambda$ & F1 & TVD & JSD \\
            \midrule
            \multirow{11}{*}{Anemia} & 0 & 0.936 & 0.008 & 0.056 \\
             & 1e-4 & 0.937 & 0.009 & 0.090 \\
             & 2e-4 & 0.938 & 0.010 & 0.194 \\
             & 3.5e-4 & 0.936 & 0.014 & 0.387 \\
             & 5e-4 & 0.942 & 0.017 & 0.481 \\
             & 0.001 & 0.938 & 0.030 & 0.576 \\
             & 0.002 & 0.895 & 0.068 & 0.666 \\
             & 0.004 & 0.888 & 0.074 & 0.690 \\
             & 0.005 & 0.875 & 0.079 & 0.692 \\
             & 0.01 & 0.872 & 0.086 & 0.693 \\
             \\
            \midrule
            \multirow{16}{*}{SST} & 0 & 0.816 & 0.032 & 0.075 \\
             & 1e-5 & 0.822 & 0.030 & 0.042 \\
             & 5e-5 & 0.824 & 0.031 & 0.080 \\
             & 1e-4 & 0.823 & 0.032 & 0.064 \\
             & 5e-4 & 0.828 & 0.031 & 0.100 \\
             & 5.2e-4 & 0.827 & 0.036 & 0.150 \\
             & 5.25e-4 & 0.823 & 0.036 & 0.514 \\
             & 5.35e-4 & 0.814 & 0.039 & 0.420 \\
             & 5.5e-4 & 0.809 & 0.040 & 0.505 \\
             & 6e-4 & 0.813 & 0.039 & 0.513 \\
             & 7.5e-4 & 0.811 & 0.037 & 0.518 \\
             & 0.001 & 0.815 & 0.038 & 0.623 \\
             & 0.01 & 0.821 & 0.036 & 0.624 \\
             & 0.1 & 0.811 & 0.047 & 0.653 \\
             & 0.5 & 0.799 & 0.061 & 0.652 \\
             & 1 & 0.819 & 0.039 & 0.624 \\
        \end{tabular}
        &
        \begin{tabular}{lcccc}
            Dataset & $\lambda$ & F1 & TVD & JSD \\
            \midrule
            \multirow{11}{*}{Diabetes} & 0 & 0.779 & 0.012 & 0.098 \\
             & 1e-5 & 0.769 & 0.011 & 0.098 \\
             & 2e-5 & 0.770 & 0.011 & 0.098 \\
             & 4e-5 & 0.780 & 0.012 & 0.162 \\
             & 5e-5 & 0.781 & 0.012 & 0.209 \\
             & 1e-4 & 0.781 & 0.016 & 0.385 \\
             & 2e-4 & 0.775 & 0.015 & 0.409 \\
             & 5e-4 & 0.759 & 0.029 & 0.494 \\
             & 0.001 & 0.690 & 0.051 & 0.646 \\
             & 0.005 & 0.645 & 0.067 & 0.693 \\
             & 0.01 & 0.643 & 0.069 & 0.693 \\
            \midrule
            \multirow{16}{*}{IMDb} & 0 & 0.906 & 0.011 & 0.043 \\
             & 1e-4 & 0.907 & 0.011 & 0.027 \\
             & 2e-4 & 0.906 & 0.012 & 0.059 \\
             & 4e-4 & 0.906 & 0.015 & 0.202 \\
             & 5e-4 & 0.905 & 0.016 & 0.141 \\
             & 7e-4 & 0.902 & 0.015 & 0.309 \\
             & 8e-4 & 0.906 & 0.014 & 0.405 \\
             & 0.001 & 0.905 & 0.022 & 0.615 \\
             & 0.005 & 0.888 & 0.038 & 0.691 \\
             & 0.01 & 0.885 & 0.039 & 0.691 \\
             \\
             \\
             \\
             \\
             \\
             \\
        \end{tabular}
        \\
        \bottomrule
    \end{tabular}
    \caption{All results for the Adversarial Setup. $\lambda=0$ denotes a model where only minimum $\TVD$ is sought. SST models were trained for 80 epochs with best epoch selected based on loss objective over the test set; all other models for 40 epochs.}
    \label{tab:all_advs}
\end{table*}

\autoref{tab:all_advs} presents results for all $\lambda$ settings tested for the trained adversary experiment performed in \S \ref{sec:oppose}.
No post-selection was made, as these are all test set results.

%% file: 94_class_plots.tex
\section{TVD/JSD Tradeoff by Class}
\label{app:class_plots}

\begin{figure}
    \centering
    \includegraphics[width=7.7cm]{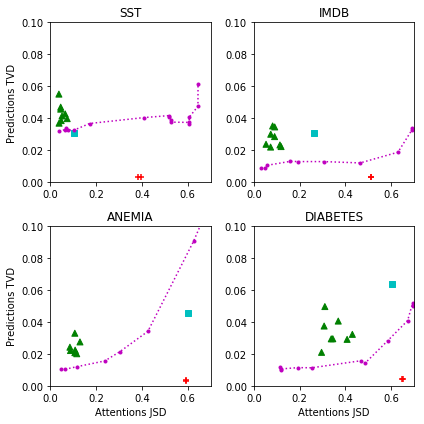} \\
    \hrule
    \includegraphics[width=7.7cm]{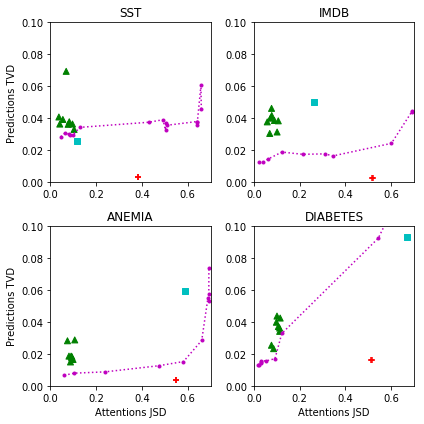}
    \caption{Per-instance test set JSD and TVD from base model on negative instances (top) and positive instances (bottom). \textcolor{olive}{$\blacktriangle$}: random seed; \textcolor{cyan}{$\blacksquare$}: uniform weights; dotted line: our adversarial setup; \textcolor{red}{$\boldsymbol{+}$}: adversarial setup from \newcite{jain2019attention}.}
    \label{fig:classplots}
\end{figure}

\autoref{fig:classplots} breaks down the scatterplots from \autoref{fig:jsdtvd} into those pertaining to each data class.
We note the unique, near-concave shape of the positive class in the Diabetes dataset, which is a detection-type set biased towards the negative class.
This is the setting where we would expect adversarial distributions to be most difficult to find, which is confirmed by this curve.

%% file: 00_attention_might.bbl
\begin{thebibliography}{19}
\expandafter\ifx\csname natexlab\endcsname\relax\def\natexlab#1{#1}\fi

\bibitem[{Bahdanau et~al.(2014)Bahdanau, Cho, and Bengio}]{bahdanau2014neural}
Dzmitry Bahdanau, Kyunghyun Cho, and Yoshua Bengio. 2014.
\newblock Neural machine translation by jointly learning to align and
  translate.
\newblock \emph{arXiv preprint arXiv:1409.0473}.

\bibitem[{Doshi-Velez and Kim(2017)}]{doshi2017towards}
Finale Doshi-Velez and Been Kim. 2017.
\newblock Towards a rigorous science of interpretable machine learning.
\newblock \emph{arXiv preprint arXiv:1702.08608}.

\bibitem[{Ehsan et~al.(2019)Ehsan, Tambwekar, Chan, Harrison, and
  Riedl}]{ehsan2019automated}
Upol Ehsan, Pradyumna Tambwekar, Larry Chan, Brent Harrison, and Mark~O Riedl.
  2019.
\newblock Automated rationale generation: a technique for explainable ai and
  its effects on human perceptions.
\newblock In \emph{Proceedings of the 24th International Conference on
  Intelligent User Interfaces}, pages 263--274. ACM.

\bibitem[{Hochreiter and Schmidhuber(1997)}]{hochreiter1997long}
Sepp Hochreiter and J{\"u}rgen Schmidhuber. 1997.
\newblock Long short-term memory.
\newblock \emph{Neural computation}, 9(8):1735--1780.

\bibitem[{Jain and Wallace(2019)}]{jain2019attention}
Sarthak Jain and Byron~C. Wallace. 2019.
\newblock {Attention is not Explanation}.
\newblock In \emph{Proceedings of the 2019 Conference of the North American
  Chapter of the Association for Computational Linguistics: Human Language
  Technologies, Volume 1 (Long Papers)}, Minneapolis, Minnesota. Association
  for Computational Linguistics.

\bibitem[{Johnson et~al.(2016)Johnson, Pollard, Shen, Li-wei, Feng, Ghassemi,
  Moody, Szolovits, Celi, and Mark}]{johnson2016mimic}
Alistair~EW Johnson, Tom~J Pollard, Lu~Shen, H~Lehman Li-wei, Mengling Feng,
  Mohammad Ghassemi, Benjamin Moody, Peter Szolovits, Leo~Anthony Celi, and
  Roger~G Mark. 2016.
\newblock Mimic-iii, a freely accessible critical care database.
\newblock \emph{Scientific data}, 3:160035.

\bibitem[{Lei et~al.(2016)Lei, Barzilay, and Jaakkola}]{lei2016rationalizing}
Tao Lei, Regina Barzilay, and Tommi Jaakkola. 2016.
\newblock \href {https://doi.org/10.18653/v1/D16-1011} {Rationalizing neural
  predictions}.
\newblock In \emph{Proceedings of the 2016 Conference on Empirical Methods in
  Natural Language Processing}, pages 107--117, Austin, Texas. Association for
  Computational Linguistics.

\bibitem[{Lipton(2016)}]{lipton2016mythos}
Zachary~C Lipton. 2016.
\newblock \href {https://arxiv.org/abs/1606.03490} {The mythos of model
  interpretability}.
\newblock \emph{arXiv preprint arXiv:1606.03490}.

\bibitem[{Maas et~al.(2011)Maas, Daly, Pham, Huang, Ng, and
  Potts}]{maas2011learning}
Andrew~L Maas, Raymond~E Daly, Peter~T Pham, Dan Huang, Andrew~Y Ng, and
  Christopher Potts. 2011.
\newblock Learning word vectors for sentiment analysis.
\newblock In \emph{Proceedings of the 49th annual meeting of the association
  for computational linguistics: Human language technologies-volume 1}, pages
  142--150. Association for Computational Linguistics.

\bibitem[{Mullenbach et~al.(2018)Mullenbach, Wiegreffe, Duke, Sun, and
  Eisenstein}]{mullenbach2018explainable}
James Mullenbach, Sarah Wiegreffe, Jon Duke, Jimeng Sun, and Jacob Eisenstein.
  2018.
\newblock \href {https://doi.org/10.18653/v1/N18-1100} {Explainable prediction
  of medical codes from clinical text}.
\newblock In \emph{Proceedings of the 2018 Conference of the North American
  Chapter of the Association for Computational Linguistics: Human Language
  Technologies, Volume 1 (Long Papers)}, pages 1101--1111, New Orleans,
  Louisiana. Association for Computational Linguistics.

\bibitem[{Nikfarjam et~al.(2015)Nikfarjam, Sarker, O'connor, Ginn, and
  Gonzalez}]{nikfarjam2015pharmacovigilance}
Azadeh Nikfarjam, Abeed Sarker, Karen O'connor, Rachel Ginn, and Graciela
  Gonzalez. 2015.
\newblock Pharmacovigilance from social media: mining adverse drug reaction
  mentions using sequence labeling with word embedding cluster features.
\newblock \emph{Journal of the American Medical Informatics Association},
  22(3):671--681.

\bibitem[{Riedl(2019)}]{riedl2019human}
Mark~O Riedl. 2019.
\newblock Human-centered artificial intelligence and machine learning.
\newblock \emph{Human Behavior and Emerging Technologies}, 1(1):33--36.

\bibitem[{Rockt{\"a}schel et~al.(2015)Rockt{\"a}schel, Grefenstette, Hermann,
  Ko{\v{c}}isk{\`y}, and Blunsom}]{rocktaschel2015reasoning}
Tim Rockt{\"a}schel, Edward Grefenstette, Karl~Moritz Hermann, Tom{\'a}{\v{s}}
  Ko{\v{c}}isk{\`y}, and Phil Blunsom. 2015.
\newblock Reasoning about entailment with neural attention.
\newblock \emph{arXiv preprint arXiv:1509.06664}.

\bibitem[{Ross et~al.(2017)Ross, Hughes, and Doshi-Velez}]{ross2017right}
Andrew~Slavin Ross, Michael~C Hughes, and Finale Doshi-Velez. 2017.
\newblock Right for the right reasons: training differentiable models by
  constraining their explanations.
\newblock In \emph{Proceedings of the 26th International Joint Conference on
  Artificial Intelligence}, pages 2662--2670. AAAI Press.

\bibitem[{Rudin(2018)}]{rudin2018please}
Cynthia Rudin. 2018.
\newblock Please stop explaining black box models for high stakes decisions.
\newblock \emph{arXiv preprint arXiv:1811.10154}.

\bibitem[{Serrano and Smith(2019)}]{serrano2019attention}
Sofia Serrano and Noah~A. Smith. 2019.
\newblock \href {https://www.aclweb.org/anthology/P19-1282} {Is attention
  interpretable?}
\newblock In \emph{Proceedings of the 57th Annual Meeting of the Association
  for Computational Linguistics}, pages 2931--2951, Florence, Italy.
  Association for Computational Linguistics.

\bibitem[{Socher et~al.(2013)Socher, Perelygin, Wu, Chuang, Manning, Ng, and
  Potts}]{socher2013recursive}
Richard Socher, Alex Perelygin, Jean Wu, Jason Chuang, Christopher~D Manning,
  Andrew Ng, and Christopher Potts. 2013.
\newblock Recursive deep models for semantic compositionality over a sentiment
  treebank.
\newblock In \emph{Proceedings of the 2013 conference on empirical methods in
  natural language processing}, pages 1631--1642.

\bibitem[{Thorne et~al.(2019)Thorne, Vlachos, Christodoulopoulos, and
  Mittal}]{thorne2019generating}
James Thorne, Andreas Vlachos, Christos Christodoulopoulos, and Arpit Mittal.
  2019.
\newblock \href {https://doi.org/10.18653/v1/N19-1101} {Generating token-level
  explanations for natural language inference}.
\newblock In \emph{Proceedings of the 2019 Conference of the North {A}merican
  Chapter of the Association for Computational Linguistics: Human Language
  Technologies, Volume 1 (Long and Short Papers)}, pages 963--969, Minneapolis,
  Minnesota. Association for Computational Linguistics.

\bibitem[{Xu et~al.(2015)Xu, Ba, Kiros, Cho, Courville, Salakhudinov, Zemel,
  and Bengio}]{xu2015show}
Kelvin Xu, Jimmy Ba, Ryan Kiros, Kyunghyun Cho, Aaron Courville, Ruslan
  Salakhudinov, Rich Zemel, and Yoshua Bengio. 2015.
\newblock Show, attend and tell: Neural image caption generation with visual
  attention.
\newblock In \emph{International conference on machine learning}, pages
  2048--2057.

\end{thebibliography}
